\documentclass[10pt,twocolumn]{article}

\usepackage{times}
\usepackage{epsfig}

\usepackage{amsmath}
\usepackage{amssymb}
\usepackage{acronym}

\usepackage{graphicx}
\usepackage{caption}
\usepackage{subcaption}

\newcommand{\Section}[1]{\vspace{-8pt}\section{\hskip -1em.~~#1}\vspace{-3pt}} 
\newcommand{\SubSection}[1]{\vspace{-3pt}\subsection{\hskip -1em.~~#1}
     	\vspace{-3pt}}

\begin{document}

\acrodef{MRF}{Markov Random Field}
\acrodef{MCMC}{Markov Chain Monte Carlo}
\acrodef{MH}{Metropolis-Hastings}

\date{}
\title{\Large\bf Multiple target tracking with interaction using an MCMC MRF Particle Filter}


\author{\begin{tabular}[t]{c@{\extracolsep{8em}}c} 
Helder F. S. Campos  & Nuno M. C. Paulino \\
 \\
        DEEC & DEEC\\
        Faculdade de Engenharia &  Faculdade de Engenharia\\
        Universidade do Porto & Universidade do Porto\\
        Rua Dr. Roberto Frias, s/n & Rua Dr. Roberto Frias, s/n\\ 
       4200-465 Porto, Portugal & 4200-465 Porto, Portugal\\
       helder.campos@fe.up.pt & nmcp@fe.up.pt
\end{tabular}}
\maketitle

\maketitle

\section*{\centering Abstract}
{\em
This paper presents and discusses an implementation of a multiple target tracking method, which is able to deal with target interactions and prevent tracker failures due to hijacking. The referenced approach uses a \acf{MCMC} sampling step to evaluate the filter and constructs an efficient proposal density to generate new samples. This density integrates target interaction terms based on \acfp{MRF} generated per time step. The \acp{MRF} model the interactions between targets in an attempt to reduce tracking ambiguity that typical particle filters suffer from when tracking multiple targets. A test sequence of 662 grayscale frames containing 20 interacting ants in a confined space was used to test both the proposed approach and a set of importance sampling based independent particle filters, to establish a performance comparison. It is shown that the implemented approach of modeling target interactions using \ac{MRF} successfully corrects many of the tracking errors made by the independent, interaction unaware, particle filters.
}

\Section{Introduction}

This report presents the implementation and results of an interaction enabled tracker approach~\cite{Khan03tr}. The work is concerned with the 
problem
of tracking multiple interacting objects such as ants or other types
of interacting targets. The problem of the usual approaches, such
as using an independent particle filter for each target, is that when
the targets become too close, the trackers become hijacked and usually one
of three effects occurs: two trackers start
tracking the same object, the trackers switch between objects or
the interaction does not influence the tracking result.
Obviously this uncertainty lowers the robustness of the trackers so new approaches are needed.

The usual approach to solve this problem is to assume that each object
will not change its motion behavior abruptly, such as it is done
with Kalman filters~\cite{kalman} by the use of an appropriate motion
model. The animal tracking problem, however, is not enhanced with such
assumption since the presence of other objects or animals will usually
interfere with the animal behavior. For example, in the test sequence here presented, when an ant encounters
another on its path, it will not continue with the same speed, such
as the usual approaches assume, but will instead stop and most of the times
reverse its direction as a result of the interaction. 

The approach to the problem made in~\cite{Khan03tr} tries to address the issue
by augmenting the motion model with additional complexity reflecting interaction between targets. Their work
is based on the well known particle filters which approximate the 
ideal Bayesian filter using Monte-Carlo methods~\cite{Pfilters}. They,
however, replace the importance sampling step, usually present in
conventional particle filters, by an \ac{MCMC} sampling step~\cite{hastings1970monte}.
This is done because, to model animal interaction, filters that evaluate 
the state of the multiple objects jointly are
needed. Joint particle filters with importance sampling could be used
but their complexity grows exponentially with the number of tracked objects
which makes them unusable for more than three or four targets~\cite{Khan03tr}. 

The remainder of this report is organized as follows: Section~\ref{sec:overview} overviews the contents of the paper, details briefly on tracking approaches using particle filters and explains the proposed tracking approach; Section~\ref{sec:experiments} presents the experimental setup; Section~\ref{sec:results} presents the results and a comparison with the approach introduced in the referenced article, finally Section~\ref{sec:conclude} concludes the report.



\Section{Overview}\label{sec:overview}

In this paper, we test and present an implementation of an interaction aware particle filter presented in~\cite{Khan03tr} along with a critical review and comparison of results. The filter is evaluated at each time step by a Markov Chain Monte Carlo (MCMC) sampling step~\cite{Greenbert95}. The targets to track in the test sequence are modeled in a single joint state, and likelihoods are computed via template matching approach. To deal with interactions a Markov Random Field (MRF) is established at each time step for near-by targets. Pairwise interaction terms are computed for these neighbourhoods which are used too resolve tracker ambiguity during the sampling stage for targets in close proximity. The following subsections explain the concept of MCMC methods, the used interaction potential approach, how the two concepts are integrated into the MCMC MRF Particle Filter along with the used motion model and likelihood measure. The performance of the proposed approach is verified by comparison with a set of independent particle filters each tracking a single target with importance sampling~\cite{Isard98condensation-}, using the same test set and setup. 

\SubSection{Multiple Target Tracking}\label{sub:mtt}


A tracker can be expressed as a Bayes filter, in which the system state \begin{math}X_t\end{math} can be modeled as the position of the target to track (which may also include velocity). For multiple target tracking the system state may include the position of all \emph{n} targets \begin{math}{X_t = \{X_{it}|i \in 1..n\}}\end{math}. Tracking is performed based on the assumption that if we know the state transition
probability density $p(X_t|X_{t-1})$ and the old posterior density
$p(X_{t-1}|Z_{t-1})$ one can predict the next state density $p(X_t|Z_{t-1})$
using the following equation:
\begin{equation}
	p(X_t|Z_{t-1}) = \int p(X_{t}|X_{t-1}) p(X_{t-1}|Z_{t-1})\  
			\mathrm{d}X_{t-1}
\label{eq:estimate}
\end{equation}
Now, using the Bayes theorem, one can calculate the posterior density
of state $X_t$ knowing the measures $Z_t$ by:
\begin{equation*}
	p(X_t|Z_t) = kp(Z_t|X_t)p(X_t|Z_{t-1})
\end{equation*}
where $k$ is a normalizing constant and $p(Z_t|X_t)$ is the likelihood
of the measurement $Z_t$ given the known state $X_t$.

The problem with this approach is that we cannot evaluate equation~(\ref{eq:estimate}) for all $X_{t-1}$. Taking the ant tracking problem as
an example, one would need to have an estimate of $p(X_{t-1}|Z_{t-1})$ for
all the coordinates of the tracking space and evaluate $p(X_t|X_{t-1})$ for all of them as well which
is practically not feasible.

Particle filters are well known approaches that approximate this Bayesian filter using Monte-Carlo methods. They are based on generating hypothesis for a given state of the system being modeled, applying a state transition model at each time step and then resampling the generated new states according to their respective likelihoods, assuming a known likelihood measure for the hypothesis. Tracking can thus be performed given an \emph{M} number of particles (state hypothesis) per time step. The rationale here is that if the prior
density $p(X_{t-1}|Z_{t-1})$ is sampled to generate $M$ samples
$s_i, i=[1 .. M]$ one can approximate the moment $g(X_t)$
by,
\begin{equation*}
	E[g(X_t)] = \sum_{n=1}^{M} g(s_i)\pi_i
\end{equation*}
with:
\begin{equation*}
	\pi_i = \frac{p(Z_t|s_i)}{\sum_{j=1}^M p(Z_t|s_j)}
\end{equation*}
thus effectively modeling $p(X_t|Z_t)$ as
\begin{equation*}
	p(X_t|Z_t) = \sum_{i=1}^{M}\pi_i\delta(X_t - s_i)
\end{equation*}
or, in other words, modeling the posterior distribution as a set of $M$ weighted samples~\cite{Isard98condensation-}.

It is assumed that the state transition probability, or motion model, for a tracker with \emph{n} targets can be factored as:
\begin{equation}
	p(X_t|X_{t-1}) = \prod_{i}^{n}p(X_{it}|X_{i, (t-1)})
	\label{eq:factoredMM}
\end{equation}
This factorization reflects the notion of a interaction unaware motion model, i.e.\ each target does not influence the remaining targets behaviour (transition), regardless of the relationship of their states \begin{math}X_{it}\end{math}.

According to~\cite{Khan03tr}, in order to augment the model with some
kind of target interaction, one needs to jointly estimate the state of all
targets. A joint particle filter in which the state is \begin{math}{X_t = \{X_{it}|i \in 1..n\}}\end{math} can be used to track \emph{n} targets. However, this increased complexity in state
space makes the filter exponentially complex with the number of targets.
To go around such complexity, fewer particles could be used but this would result in undersampling of the state space, making the filter perform poorly. For this reason, a MCMC sampling step is used in the referenced approach
so that the low complexity of independent particle filters, augmented with an interaction model, could be used.


\SubSection{MCMC Sampling}\label{sub:MCMCsample}

In order to sample from potentially non-trivial posterior distributions, more efficient \acf{MCMC} sampling methods can be used, instead of the sequential importance sampling approach of particle filters. Such sampling methods rely on generating a random walk over a target distribution, in this case the posterior $p(X_t|Z_t)$. The Markov property of the process comes from the fact that the next sample is generated in function of the last generated sample. Given a large enough number of iterations, the chain of samples will converge to the target distribution. The approximation accuracy increases with both the number of samples and iterations.

As with any Markov process, this sampling method requires a transition model, which in this case will generate new sample proposals from the current sample. This proposal density, \begin{math}q(x, y)\end{math}, must satisfy a reversibility condition which states that the transition probability from a given state \begin{math}x\end{math} to a state \begin{math}y\end{math} must equal the probability of transitioning from \begin{math}y\end{math} to \begin{math}x\end{math}. The \ac{MH} algorithm is able to assure this condition~\cite{Greenbert95}.

To generate a Markov chain of samples using the MH algorithm a valid starting sample, \begin{math}x^{j}\end{math}, must be chosen. To generate \emph{N} new samples the following steps are repeated:
\begin{enumerate}
	\item{Generate a sample proposal \begin{math}x{'}^{j+1}\end{math} using \begin{math}q(x^{j}, ~.)\end{math}}
	\item{Generate a number \emph{u} from the uniform distribution}
	\item{If \begin{math}u < \alpha(x^{j}, x{'}^{j+1})\end{math} set \begin{math}x^{j+1} = x{'}^{j+1}\end{math}}
	\item{Else set \begin{math}x^{j+1} = x^{j}\end{math}}
\end{enumerate}

New sample hypothesis are accepted or rejected according to the acceptance ratio, \begin{math}\alpha(x, y)\end{math}, which the MH algorithm introduces to assure reversibility. The derivations are out of the scope of this paper. Suffice it to say that \begin{math}\alpha(x, y)\end{math} is defined as show in in~(\ref{eq:acceptRatio1}).
\begin{equation}
\alpha(x, y) = min\bigg[1, \frac{\pi(y)q(y, x)}{\pi(x)q(x, y)}\bigg]
\label{eq:acceptRatio1}
\end{equation}

Where \begin{math}\pi\end{math} is the target distribution being sampled, in this case the posterior probability of the particle filter. Section~\ref{sub:MCMCMRF} details on the proposal density used for this implementation, which includes the interaction terms explained in the following section.

\SubSection{Modeling Target Interactions}\label{sub:interact}

Interactions occur between tracking targets when their relative proximity influences their behaviour, thus invalidating the motion models that fail to capture this influence. That is, independent trackers might shift between the target they were tracking to a near-by target, if its likelihood overwhelms the likelihood of the current target. This occurs when only a small number of particles drifts (due to the motion model) towards a nearby target of higher likelihood. In the Metropolis-Hastings sampler for instance, this is expressed as an acceptance ratio superior to 1, which means the new proposal's likelihood is superior to the previously generated sample. For a sequential importance sampler, the weights of these few particles will overwhelm the weights of the more numerous particles still tracking the correct target, which will cause a tracker hijack.

To avoid switching of targets, the referenced approach aims to resolve these interaction conflicts by introducing an interaction term to the acceptance ratio that counteracts the effect of high likelihoods that appear when evaluating new samples which drift onto proximal targets.

To do so, the motion model \begin{math}P(X_{it}|X_{i(t-1)})\end{math} is augmented with interaction terms computed on-the-fly, per time step and for targets within a neighbourhood defined by a \acf{MRF}. A \ac{MRF} is a graph type representation of a set of random variables that have the Markov property, i.e.\ their next state depends only on their current state. Variables are represented by a set of nodes connected by undirected edges that represent the influence nodes impose on each others Markov processes. For a \ac{MRF}, the joint probability of the variables within the field can be calculated as a product of terms named \emph{factor potentials}. For this application, \ac{MRF} nodes are the tracking targets, and edges represent interactions. Targets are modeled as interacting in a pairwise fashion within their neighbourhood. One interaction term between two targets constitutes one factor potential, \begin{math}\psi\end{math}, of the \ac{MRF}. 

 So, for a given neighbourhood \begin{math}E\end{math} the joint probability of a given node (target) configuration is given by~(\ref{eq:iPot1}).

\begin{equation}
\prod_{ij \in E}\psi(X_{it}, X_{jt})
\label{eq:iPot1}
\end{equation}

For the purpose of multiple target tracking in the example application we present here, what is desired is a joint probability that evaluates node configurations in which nodes share the same state as unlikely. That is, a target may not share the same state (position) as another target. In other words, if any one target \emph{i} in \begin{math}X_{t}\end{math} overlaps the position of another target \emph{j} within its neighbourhood, the end product will be smaller, thus reducing the probability given by the motion model. This expresses the notion that no two, or more, targets may occupy the same space. Coupling this joint probability to the factored motion model~(\ref{eq:factoredMM}) of all targets as an additional weighting term results in the MRF motion model equation given by~(\ref{eq:MRFmotionmodel}).
\begin{equation}
P(X_{t}|X_{t-1}) \propto \prod_{i}P(X_{it}|X_{i(t-1)}) \prod_{ij \in E}\psi(X_{it}, X_{jt})
\label{eq:MRFmotionmodel}
\end{equation}

A new sample derived from the factored motion model that decreases the pairwise interaction potentials between targets is automatically weighted down due to the low joint node configuration probability given by the \ac{MRF}. Equation~(\ref{eq:MRFmotionmodel}) however expresses a proportionally relationship and not an equality. For a valid model, the two distributions need be well normalized so the \ac{MRF} motion model has the desired behaviour. The following section shows how integrating this motion model with the \ac{MCMC} sampling step removes the need to find any normalizing constants and arrives at a straightforward sampling algorithm that is interaction aware.


\SubSection{MCMC MRF Particle Filter}\label{sub:MCMCMRF}

It can be shown that plugging the interaction aware motion model~(\ref{eq:MRFmotionmodel}) into a unweighted sample approximation to the posterior analogous to that of the weighted particle filter we can arrive at the posterior approximation given by the following equation:
\begin{equation}
\begin{split}
P(X_t|Z_t)	\approx 	& kP(Z_t|X_t)\\
	 			& \prod_{ij \in E} \psi{X_{it}, X_{jt}} \sum_{r} \prod_{i}P(X_{it}|X^{(r)}_{i(t-1)})
\end{split}
\label{eq:MRFposterior}
\end{equation}

Where the interaction terms $\psi$ are not dependent on $r$ and so they can be moved out and act as another multiplicative factor along with the likelihood $P(Z_t|X_t)$. Instead of importance sampling, this posterior is now sampled using a \ac{MCMC} step, which is more efficient in spaces of high dimensionality.

As shown before in equation~(\ref{eq:acceptRatio1}), in order to generate and evaluate samples using an \ac{MCMC} sampler we require a proposal density \begin{math}q(x, y)\end{math} and must be able to evaluate the posterior at the generated sample. The \ac{MH} algorithm assures that the proposal is reversible (i.e.\ \begin{math}q(x, y) = q(y, x)\end{math}) due to its criteria of accepting samples with a probability given by the acceptance ratio~(\ref{eq:acceptRatio1}). So, any proposal density can be used. The authors in~\cite{Khan03tr} construct a proposal that is based on applying the motion model to one target at a time, per each \ac{MH} sampling step:
\begin{equation}
Q(X^{'}_{t}, X_{t}) = P(X^{'}_{it}|X_{it})
\label{eq:proposal1}
\end{equation}

Where $X^{'}_{t}$ is the new sample proposal and $X_{t}$ the previously generated sample. By plugging the \ac{MRF} posterior distribution~(\ref{eq:MRFposterior}) and the proposal density~(\ref{eq:proposal1}) into the \acf{MH} acceptance ratio, and making use of the Bayes Theorem, we arrive at the criteria used for evaluating new samples used by the referenced approach:
\begin{equation}
as = min\bigg[1, \frac{P(Z_t|X^{'}_{it})\prod_{j \in E_i}\psi(X^{'}_{it}, X^{'}_{jt})} {P(Z_t|X_{it})\prod_{j \in E_i}\psi(X_{it}, X_{jt})}\bigg]
\label{eq:acceptRatioMCMC}
\end{equation}
It is thus shown that we do not require knowing the normalization terms present in the \ac{MRF} motion model~(\ref{eq:MRFmotionmodel}) in order to include the interaction terms $\psi$ in the sampling procedure, only a likelihood measure.

The employed \ac{MCMC} based sampling algorithm can be summarized by the following steps:
\begin{enumerate}
	\item{Start with a set of joint particles for time \emph{t-1} \begin{math}\{X^{(r)}_{t-1}\}^{N}_{r=1}\end{math}}

	\item{Start the MCMC sampler by applying the factored motion model to a randomly selected particle \begin{math}X^{(r)}_{t-1}\end{math}. This generates the starting sample \begin{math}X_{t}\end{math}.}

	\item{Perform \ac{MH} iterations to generate \emph{N} samples from the posterior~(\ref{eq:MRFposterior}):}
		\subitem{- Select a random joint particle \begin{math}X^{(r)}_{t-1}\end{math}}
		\subitem{- Select a random target \emph{i} in that particle}
		\subitem{- Attain a new proposal from \begin{math}P(X_{it}|X^{(r)}_{i(t-1)})\end{math}}
		\subitem{- Evaluate the acceptance ratio~(\ref{eq:acceptRatioMCMC})}
		\subitem{- Set the current \begin{math}X_{it}\end{math} to \begin{math}X^{'}_{it}\end{math} if accepted}
		\subitem{- Add the current \begin{math}X_{t}\end{math} to the sample set}

	\item{\begin{math}\{X^{(s)}_{t}\}^{N}_{s=1}\end{math} is the new sample set for time \emph{t}}
\end{enumerate}
	
\Section{Experimental Setup}\label{sec:experiments}



To test the performance of the MCMC MRF particle filter a video sequence of randomly moving ants was used. The sequence consists 662 grayscale frames which are 720 by 480 pixels (taken from a sequence of 10400 frames). Frames contain 20 ants moving within a well bounded rectangular space, in a topdown viewpoint, which is about 15 by 10 centimeters. The sequence was also ran with a set of independent non interaction aware  particle filters using the CONDENSATION algorithm~\cite{Isard98condensation-}, each tracking a single ant, in order to establish a comparison to the MCMC MRF filter.

We extracted groundtruth for the sequence manually for all frames by following the position of each ant with a cursor, and storing the position of the cursor per frame. We followed each ant twice, with each run following a different end of the ant. From both tracks we extracted a mean position and the two tracks also allowed us to extract orientation of each ant. Once a tracker in either of the tracking approaches was lost or switched targets, we registered a tracker failure and automatically corrected the position of the tracker for that target. Like the referenced approach, we considered a distance of 50 pixels between a tracker and its groundtruth a tracking failure. We do not distinguish between tracking failures due to switched targets and tracking failures due to lost targets. The former occur due to interaction and the latter if the particle variance is to small to accommodate target movement speed.

The tracking state includes position and orientation. The interaction free motion model $P(X_t|X_{t-1})$ used applies motion to all targets in a given particle along their current orientation in a normal distribution along both orthogonal axes:
\begin{equation*}
\theta = \theta_{i} +  \Delta\theta
\end{equation*}
\begin{equation}
X_{it} = 
\left[
 \begin{matrix}
  cos(\theta) & -sin(\theta) & 0 \\
  sin(\theta) & cos(\theta) & 0 \\
  0 & 0 & 1
 \end{matrix}
\right]
\times
\left[
\begin{matrix}
  \Delta x \\
  \Delta y \\
  \Delta\theta
 \end{matrix}
\right]
+X_{i(t-1)}
\label{eq:mmodel}
\end{equation}

Where $[\Delta x, \Delta y, \Delta\theta]$ are the samples given from normal distributions with zero mean and $(\sigma_x, \sigma_y, \sigma_\theta) = (5.0, 3.0, 0.4)$ and $\theta$ is the current orientation of target \emph{i} as given by $X_{it}$. These standard deviations are as set in the referenced \ac{MCMC} \ac{MRF} filter approach.

To evaluate the likelihood $P(Z_t|X_t)$, a template matching approach was used. Templates for the foreground (ants) and background were learned using two sets of 32 manually selected images of 32 by 10 pixels. The mean image of each set is computed and a single pixel intensity mean is derived from this mean image. A standard deviation for both templates is computed using the pixel intensity mean of each. The likelihood is given by equation~(\ref{eq:lkmetric1}).
\begin{equation}
logP(X_{it}|Z_t) = -\frac{1}{2}\frac{\|F(X_{it}) - \mu_F\|}{\sigma_F} +\frac{1}{2}\frac{\|F(X_{it}) - \mu_B\|}{\sigma_B}
\label{eq:lkmetric1}
\end{equation}

Where \begin{math}F(X_{it})\end{math} is the vector of pixels retrieved by sampling the current frame at the position and orientation given by the hypothesis \begin{math}X_{it}\end{math}, i.e.\ the position of the \begin{math}i_{th}\end{math} target in the joint particle \begin{math}X_t\end{math}.


We also verified the performance of two other likelihood measures. One computed the mean difference between the sampled target hypothesis and the template which is then evaluated in a Gibbs distribution. Another computed the cross correlation between the sample and the template. However, both measures proved too ambiguous to provide acceptable results.

The interaction potential between two targets of the same particle is computed as a function of the number of overlapping pixels between them, as shown in equation~(\ref{eq:ipot1}), where \emph{p} is the number of overlapping pixels between the image sample taken at the position given by \begin{math}X_{it}\end{math} and the sample at the position given by \begin{math}X_{jt}\end{math}. 
\begin{equation}
\psi(X_{it}, X_{jt}) = e^{-5000p}
\label{eq:ipot1}
\end{equation}
As the area of overlap between two targets increases, the interaction potential decreases, countering any increase of the acceptance ratio due to increase of likelihood. We also tested an interaction metric based on the euclidean distance between particles, which loses awareness of the current orientation of the ants and yet another that used a circular area instead of a template sized rectangle. However, there was no notable difference in performance between either metric.

\begin{figure}[t]
	\centering
               \includegraphics[width=\columnwidth]{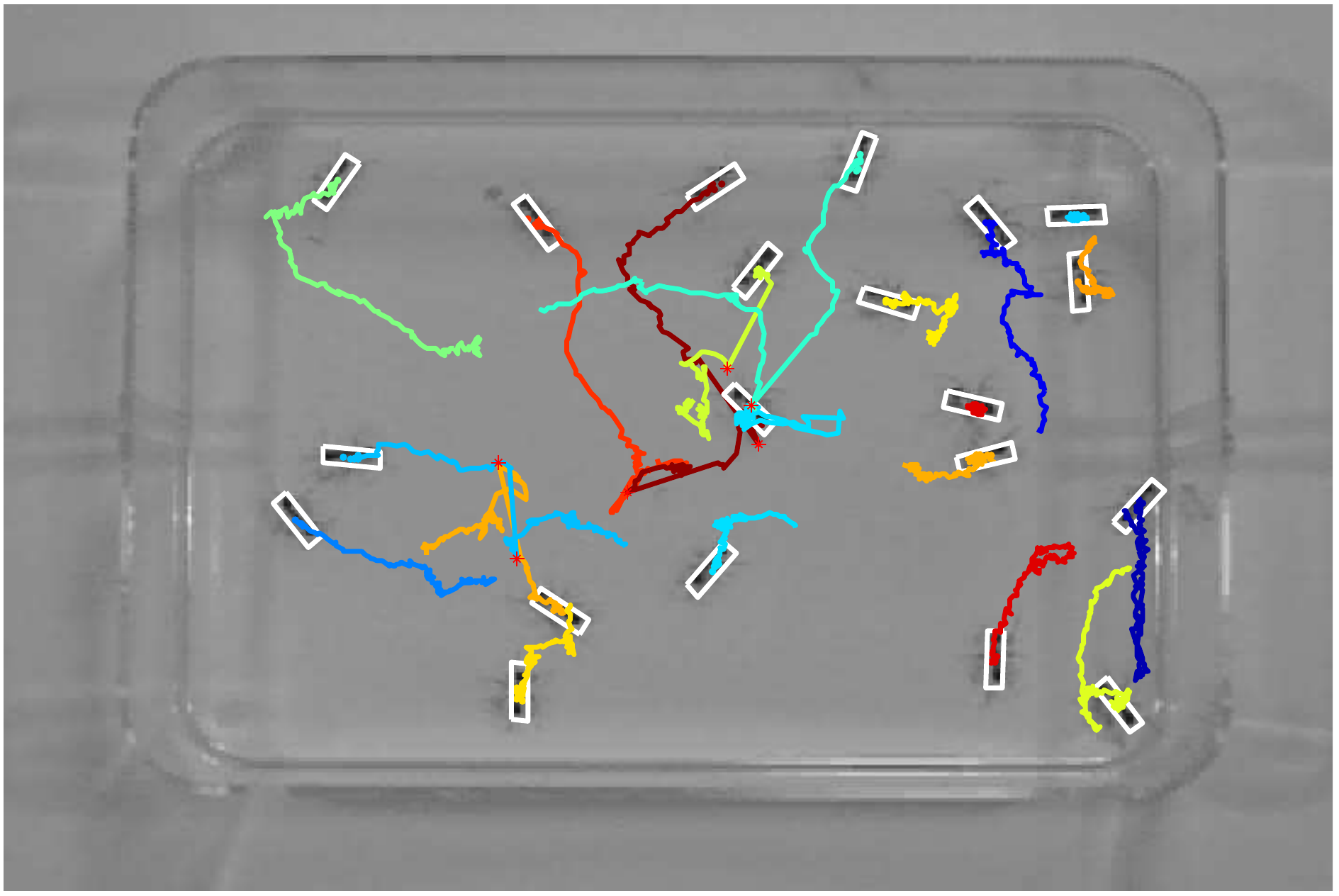}
	    \caption{Tracking status using the MCMC MRF Filter with 1000 particles, at frame 133.}
               \label{fig:ants}
\end{figure}

\section{Results}\label{sec:results}

\begin{figure}[t]
        \begin{subfigure}[t]{\columnwidth}
               \centering
               \includegraphics[width=\columnwidth]{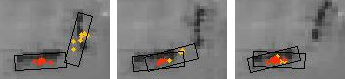}
	    \caption{Two independent particle filters suffering from target switching failure}
               \label{subfig:Indhijack}
        \end{subfigure}

        \begin{subfigure}[t]{\columnwidth}
               \centering
               \includegraphics[width=\columnwidth]{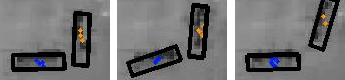}
	    \caption{Hijack avoidance due to the interaction potential counteracting the increase in likelihood}
               \label{subfig:MCMChijack}
	\end{subfigure}
	\caption{Behaviour comparison between independent and MCMC MRF filters when dealing with a relatively simple interaction}
	\label{fig:hijack}
\end{figure}

Figure~\ref{fig:ants} shows the status of the MCMC MRF tracker using 1000 particles, the likelihood metric given in~(\ref{eq:lkmetric1}) and the interaction potential term shown in~(\ref{eq:ipot1}). Red asterisks denote automatic corrections made to the trackers by the groundtruth. The colored plots denote the tracked paths of each ant. The joint particles accurately track the orientation of the ant also, represented by the white rectangles overlaid on each ant. Approximately at the center of the space, there is an intense interaction between 4 ants, and it can be seen that three tracker failures were corrected during those frames.

\begin{table*}[!t]
	\centering
	\begin{tabular}{cccc}
	\hline
	{\bf Tracker} & {\bf Number of Particles} & {\bf Number of Failures}
	& {\bf Equivalent Number of Failures} \\
	& & & {\bf (10,400 frames)} \\
	\hline
	Independent Particle Filter & 10 per target & 26 & 424 \\
	Independent Particle Filter & 50 per target & 23 & 375 \\
	Independent Particle Filter & 100 per target & 23 & 375 \\
	MCMC			& 50 & 57 & 1468 \\
	MCMC			& 100 & 29 & 473 \\
	MCMC			& 200 & 17 & 277 \\
	MCMC			& 1000 & 17 & 277 \\
	\end{tabular}	
	\caption{Tracker failure results observed in 662 frames
		from the same test sequence used 
		by~\cite{Khan03tr}.}\label{table:results}
\end{table*}

\begin{figure*}[!t]
        \begin{subfigure}[!t]{\textwidth}
               \centering
               \includegraphics[width=0.49\textwidth]{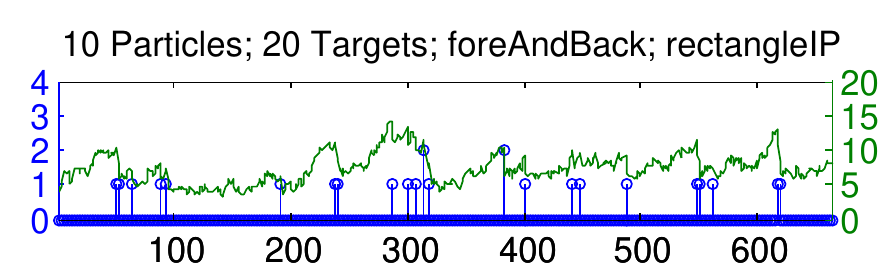}
	    \includegraphics[width=0.49\textwidth]{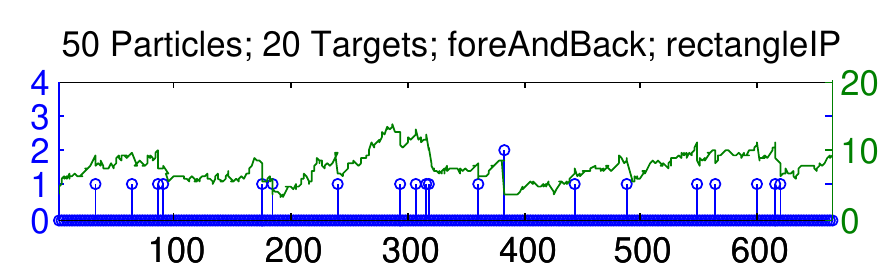}
	    \caption{Independent tracker results, using 10 and 50 particles per target. This is equivalent to 200 and 1000 particles}
               \label{subfig:Indresults}
        \end{subfigure}

        \begin{subfigure}[!t]{\textwidth}
               \centering
               \includegraphics[width=0.49\textwidth]{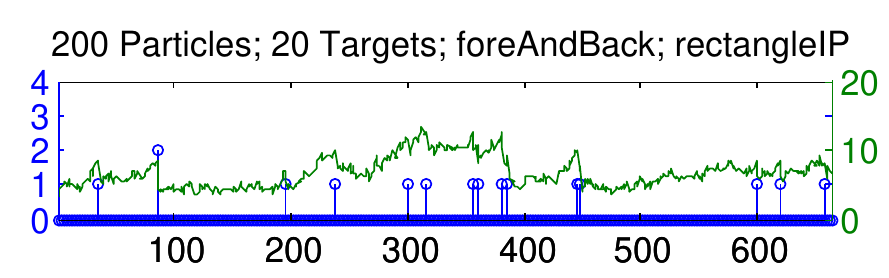}
               \includegraphics[width=0.49\textwidth]{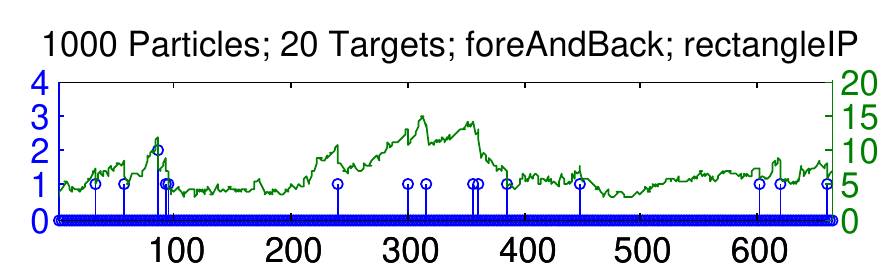}
	    \caption{MCMC MRF tracker results, using 200 and 1000 particles modeling a joint state}
               \label{subfig:MCMCresults}
	\end{subfigure}
	\caption{Tracking results for 2 runs of independent particle filters, 
		shown in~\ref{subfig:Indresults}, and MCMC tracking, show in~\ref{subfig:MCMCresults}. Horizontal axes represent frames. The left vertical axis the number of tracker failures for each frame, shown by the stems. The right vertical axis is the average distance of trackers to the groundtruth. The likelihood metric used was the one shown in equation~(\ref{eq:lkmetric1}) and the interaction potential was computed as shown in ~(\ref{eq:ipot1}).}
	\label{fig:trackresults}
\end{figure*}

Figure~\ref{fig:hijack} shows the effects of the interaction potential introduced by the MCMC MRF Filter for 3 frames of the sequence. Whereas one of the two independent trackers switches to the position of higher likelihood, i.e.\ another target, the MCMC filter weights the acceptance ratio with both the increase in likelihood and decrease in interaction potential, which discards proposal samples that would place both trackers on the same target.

Table~\ref{table:results} shows the results of 3 runs of the independent particle filters and 4 runs for the MCMC filter. In comparison to the results in~\cite{Khan03tr} our tracker implementation results in a much higher number of tracking failures. However, the decrease of failures with the increase of particles is consistent with the results reported in the referenced paper. Ten particles for the sets of independent trackers results in the same complexity as an MCMC filter with 200 joint particles. Thus, these two are comparable, as are the independent trackers  and the MCMC tracker with 50 and 1000 particles respectively. Increasing the number of particles for the independent filters does not influence the number of tracking failures as these occur due to hijacking or switched targets. The independent trackers themselves are already considerably precise, and their mean deviation from the groundtruth is comparable to that of the MCMC filters. On the other hand, increasing the number of particles for the MCMC filters does reduce the number of failures as more hypothesis help resolve the ambiguity around multiple targets, with the aid of the interaction terms.

In fact, Figure~\ref{fig:trackresults} shows 4 runs, 2 for the independent filters and 2 for the MCMC filters for the 662 frames of the sequence. The mean distance from the groundtruth for all targets is shown as a time series and the number of tracking failures as a stem series for all runs. It is noticeable that the mean distance is very similar for all tracking parameters for sequences of frames when there are no tracking failures.


\SubSection{Discussion}\label{sec:discuss}

As a comparison to the results in~\cite{Khan03tr}, we observed a few differences between our implementation and the author's reported results. More noticeably in the magnitude of tracker failures attained by the \ac{MCMC} filter. Our number of failures, for both trackers, are considerably
higher than the ones achieved by the authors, even for the same
test set. Since we have used less frames, we had to scale the number of
failures linearly to their number of frames so that a comparison could
be made. One could argue that our number of frames are not characteristic
enough to extrapolate results to a higher number of frames. However, since the authors report results along a timescale containing all frames (of the original 10400 frame sequence), we were able to establish a comparison for the number of tracker failures they achieved for that frame span, which are still consistently lower than ours. Another argument could be that, because we converted the images to
gray scale levels, information was lost explaining their improved
performance. We find this very unlikely since the images, even in
RGB space, are composed of tones of gray so the conversion did not loose
much. With this, we arrive at no possible explanation for the difference in the 
order of magnitude for the number of tracker failures, barring any unlikely implementation mistakes on our part. With the exception of the order of magnitude in the number of errors, the interaction aware tracker behaves as expected.

It is also noticable that the authors report a very low average distance to groundtruth (tracking error) for their MCMC tracker, around 1 pixel, and a considerably higher tracking error for the independent filter implementations, around 4 pixels or higher. On the other hand, our implementations of both filter types achieve very similar tracking errors for our 662 frame sequence.
The difference in mean error distance between the groundtruth
and the \ac{MCMC} tracker for ours and the referenced implementation can be explained very easily. 
The groundtruth used in the reference paper was obtained by running their own \ac{MCMC} filter with 2000 particles. 
Obviously, the error between runs of the same filter run with different particles will converge to zero
asymptotically even if they have error to the "real" ground truth. Also, creating a groundtruth like this does not establish a valid absolute  basis of comparison with others approaches, as only relative comparisons can be derived. Our approach to ground truth extraction is also not perfect since
the hand made tracks have natural human error which would make even
a perfect tracker to converge to a nonzero error. This clearly explains
the differences between error measures and, specifically, why the error
of our implementation converges to nonzero error and why theirs converges to zero. However, results concerning the reduction of tracker errors remains valid, regardless of groundtruth.

Another observation concerns the manner in which the number of interactions to deal with is presented. Authors state that the \ac{MCMC} filter deals with numerous interactions that occur consistently throughout all frames. In the manner in which they measured the number of interactions, 2 ants need only be at a distance of 64 pixels from one another to register as neighbours in an \ac{MRF}. However, this does not necessarily imply that independent trackers would result in tracker failures from all such situations. They might successfully keep tracking both targets if the particle spread is low (i.e.\ the particles are very tightly distributed around the ant) and/or the likelihoods of neighbour targets are similar.

As a final note, we used the CONDENSATION algorithm to implement our independent particle filters, but the authors do not report the implementation to which they established a comparison. We do not believe that this greatly influenced the results, for the purposes of our implementation.









\Section{Conclusions}\label{sec:conclude}

In this paper we have confirmed the improved performance of the interaction
aware tracker of \cite{Khan03tr} over independent trackers.  It is noted
that the performance of the independent tracker does not increase considerably
with the number of particles, which is explained by the nature of the errors
which are, most of the cases, due to the interaction between targets. The
performance of the interaction aware tracker increases consistently with
the number of particles, as expected, since the interactions are well modeled.
These results are good confirmation of the arguments given in the reference
paper, where interaction modeling is concerned.

\bibliography{paper}
\bibliographystyle{ieeetr}

\end{document}